\documentclass{article}
\usepackage[preprint]{colm2026_conference}
\usepackage{microtype}
\usepackage{hyperref}
\usepackage{url}
\usepackage{booktabs}
\usepackage{graphicx}
\usepackage{subcaption}
\usepackage{amsmath}
\usepackage{amssymb}
\usepackage{lineno}
\definecolor{darkblue}{rgb}{0, 0, 0.5}
\hypersetup{colorlinks=true, citecolor=darkblue, linkcolor=darkblue, urlcolor=darkblue}

\title{TerraAgent: Demonstration-Grounded Multi-Agent Reasoning for Long-Context Remote Sensing Visual Question Answering}

\author{Jingwen Huang, Yiming Fang, Hanxiao Wu, Boyuan He, Keyi Luo \\
Yunnan University of Finance and Economics, Kunming, China \\
\texttt{keyi.luo@kust.edu.cn}
}

\begin{document}
\ifcolmsubmission
\linenumbers
\fi
\maketitle

\begin{abstract}
Remote sensing visual question answering (RS-VQA) requires answers that truly depend on image evidence, yet current remote sensing multimodal large language models (RS MLLMs) are still evaluated mainly by accuracy, which cannot reveal whether a correct answer was produced without using the image. This setting introduces three challenges: demonstrations retrieved across different ground sample distances (GSD), sensors, and city textures may be poorly matched; real queries often involve long visual contexts from multi-scale tiles, multi-temporal frames, and retrieved knowledge, where evidence can be diluted and errors can propagate through multi-step reasoning; and prior-only answering remains invisible to standard metrics. We present \textbf{AVA}, a training-free framework that wraps a frozen RS MLLM with three modules. \emph{GDR} performs GSD-bucketed, scene-graph-based demonstration retrieval and filters demonstrations whose answers are unsupported by their images. \emph{MACE} decomposes each query into an evidence-typed sub-question DAG, routes sub-questions to suitable perception tools, and records verified provenance in a shared evidence ledger with bounded re-perception. \emph{CVA} audits the final answer through counterfactual perturbations of cited evidence, including ablation and swap, and computes a visual-dependency score for confidence calibration and evidence-forcing re-query. On RSVQA-LR/HR, VRSBench, and LHRS-Bench, AVA improves average accuracy by up to $2.40$ points over the strongest same-setting baselines, raises the visual-dependency score from $0.28$ to $0.49$, reduces the prior-only answer rate from $46.2\%$ to $21.7\%$, and lowers the hallucination rate from $17.1\%$ to $9.8\%$, with $2.6\times$ the cost of a single backbone forward pass and no fine-tuning.
\end{abstract}

\section{Introduction}

Remote sensing visual question answering (RS-VQA) is the interface through which earth-observation systems turn pixels into decisions: damage assessment, illegal-construction screening and urban-expansion monitoring all end with a natural-language judgement about \emph{what is there}, \emph{what changed} and \emph{what it means} \cite{lobry2020rsvqa}. In the last two years, remote sensing multimodal large language models (RS MLLMs) such as GeoChat \cite{kuckreja2023geochat}, LHRS-Bot \cite{muhtar2024lhrsbot}, RSGPT \cite{hu2023rsgpt} and SkyEyeGPT \cite{zhan2024skyeyegpt} have pushed zero-shot RS-VQA accuracy to levels that were unreachable with task-specific pipelines, mostly by scaling instruction data and domain adaptation. Yet three properties of the remote sensing setting remain unresolved, and they are precisely the ones that determine whether such a model can be deployed.

\paragraph{The central question.} A single number hides the whole story. On RSVQA-LR, a model that never sees the image and answers from the question text alone already reaches $80.90\%$ overall accuracy, whereas the full image--text model reaches $87.14\%$ \cite{wang2024rsadapter}. Most of the reported accuracy is therefore recoverable from language priors: ``is there a building?'' has a skewed answer distribution, and ``how many'' questions collapse to a handful of values. Accuracy cannot tell us whether an answer came from the image or from the question. We therefore treat \emph{visual dependency} --- the causal sensitivity of an answer to the visual evidence it cites --- as a first-class quantity, not an after-the-fact analysis.

\paragraph{Why demonstrations do not transfer.} In-context learning is the cheapest way to adapt a large vision--language model to a new task \cite{zhou2024visual}, and retrieval-augmented variants have been ported to remote sensing, either over knowledge \cite{wen2026rsrag} or over imagery at scale \cite{rajendran2026damagescope,kazoom2026rsrcc}. However, RS imagery spans three orders of magnitude of ground sample distance (GSD): a Sentinel-2 tile at $10$~m and a USGS orthophoto at $15$~cm show the same ``urban density'' question through entirely different texture and object sizes. Cosine similarity over global image embeddings confuses a low-resolution city with a high-resolution field, and once several retrieved demonstrations share an answer (six ``yes'' out of eight), the model copies that distribution instead of reading the query image. What is missing is a comparability measure that is \emph{scale-normalised} and a reject rule that removes demonstrations whose answer is not grounded in their own image.

\paragraph{Why long contexts hurt.} Real queries are answered over long visual contexts: a large scene plus multi-scale crops (a small object occupies a few dozen pixels at full extent), multi-temporal frames, and external knowledge. Once the context grows, large vision--language models systematically shift weight from visual evidence to textual cues, and 2026 evaluations of agentic systems report three recurring failure modes --- domain-toolset mismatch, unchecked error propagation, and static reasoning \cite{zhang2026uavmas,xiao2026geommbench}. A five-step chain with $5\%$ per-step error can accumulate to roughly $23\%$ end-to-end error, and no current RS agent makes ``every claim must have provenance'' a hard constraint, although planner training for long-horizon agent tasks is known to dominate local heuristics \cite{si2026goal}.

\begin{figure}[!t]
\centering
\includegraphics[width=\linewidth]{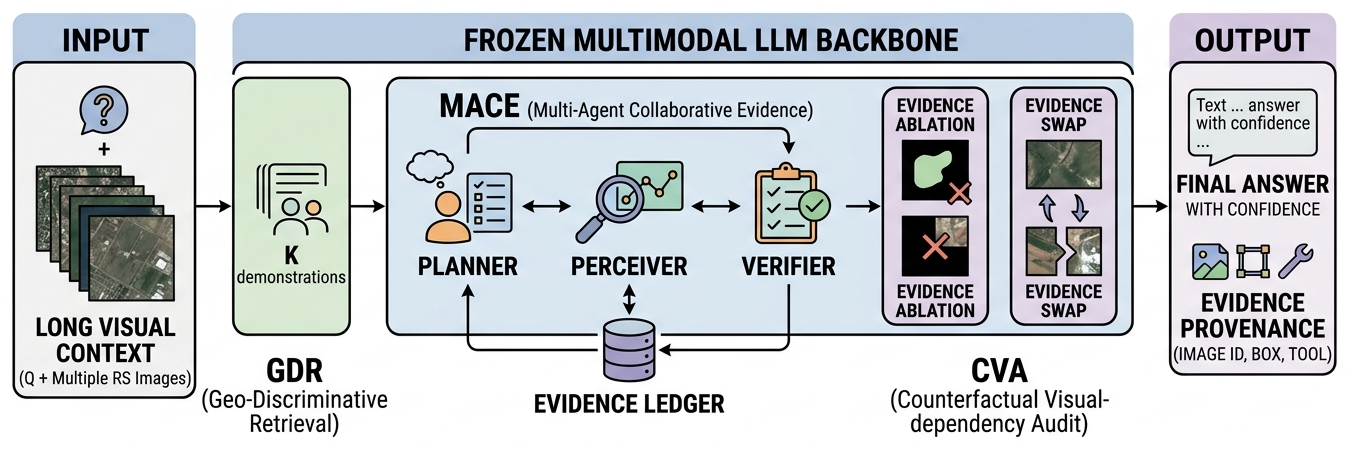}
\caption{Overview of TerraAgent. A frozen remote sensing multimodal LLM is wrapped by three training-free modules: GDR builds a scale-balanced, evidence-gated long context from retrieved demonstrations; MACE decomposes the question, routes sub-questions to perception tools, records typed evidence with provenance in a shared ledger and verifies every answer span; CVA perturbs the cited evidence (ablation and swap) to compute the visual-dependency score, which calibrates confidence and decides between acceptance, evidence-forcing re-query and abstention.}
\label{fig:arch}
\end{figure}

\paragraph{Contributions.} We present \textbf{TerraAgent} (Figure~\ref{fig:arch}), a training-free framework that equips a frozen RS MLLM with demonstration grounding, evidence-carrying collaboration, and counterfactual auditing. Our contributions are:
\begin{itemize}
\item \textbf{GDR (Geo-Discriminative Retrieval).} We replace single-vector retrieval with GSD-bucketed, dual-granularity matching (global RemoteCLIP embedding plus a structured scene graph over class counts, a $3\times3$ position histogram and a scale bucket), followed by an evidence-consistency gate (ECG) which asks the frozen backbone to re-answer each candidate demonstration from its image alone and discards candidates whose stored answer is not reproduced. The gate removes $12\%$ of the demonstration pool.
\item \textbf{MACE (Multi-Agent Collaborative Evidence).} We instantiate role-specialised collaboration in an evidence-centric form: a planner emits an evidence-typed sub-question DAG, a perceiver routes each sub-question to the task-appropriate tool, every intermediate result is written as a typed entry with provenance and confidence into a shared evidence ledger, and a verifier rejects unsupported spans and triggers bounded re-perception. The ledger turns error propagation from a statistical risk into a checkable invariant.
\item \textbf{CVA (Counterfactual Visual-dependency Audit).} We make visual dependency measurable at inference time. For each surviving answer we run two perturbations of the evidence it cites --- precise ablation of the cited region (or key temporal frame) and replacement with same-class evidence from another location --- and define the visual-dependency score (VDS) and evidence-swap sensitivity (ESS). Low-VDS answers are re-queried with evidence-forcing prompts or abstained, and VDS is used to calibrate confidence.
\item \textbf{Results and diagnosis.} On RSVQA-LR/HR \cite{lobry2020rsvqa}, VRSBench \cite{li2024vrsbench} and LHRS-Bench \cite{muhtar2024lhrsbot}, TerraAgent improves the strongest same-setting baseline by $+1.59$, $+1.55$, $+2.40$ and $+1.74$ points respectively, while raising VDS from $0.28$ to $0.49$, lowering the prior-only answer rate from $46.2\%$ to $21.7\%$ and the judged hallucination rate from $17.1\%$ to $9.8\%$. We further show that longer demonstration contexts help and then hurt (\S\ref{sec:hyper}), that the gain grows with domain shift, and that our efficiency remains $2.6\times$ a single forward pass with no training.
\end{itemize}

\section{Related Work}

\paragraph{Remote sensing visual question answering.} RSVQA established the task and the low-/high-resolution splits that are still standard \cite{lobry2020rsvqa}; VRSBench broadened evaluation to nine question categories with human-verified captions and object references \cite{li2024vrsbench}; LHRS-Bench provides an objective single-choice probe of eleven remote sensing capability dimensions with global $1$~m imagery \cite{muhtar2024lhrsbot}. Parameter-efficient adaptation of pretrained multimodal transformers remains competitive there \cite{wang2024rsadapter}. Our work keeps these benchmarks but changes what is reported: alongside accuracy we report how much of the accuracy depends on the image.

\paragraph{Remote sensing and multimodal agents.} Two 2026 systems are closest to us. GeoMMAgent orchestrates retrieval, perception and reasoning specialists over a geoscience benchmark and shows that tool-augmented agents beat standalone models \cite{xiao2026geommbench}; UAV-MAS is a training-free multi-agent system whose ablation identifies domain-toolset mismatch, unchecked error propagation and static reasoning as the dominant failure modes \cite{zhang2026uavmas}. Role-specialised collaboration was shown to outperform monolithic models earlier in the multimodal medical setting \cite{zhou2025mam}. Our MACE reuses the planning skeleton but makes the intermediate product a ledger with provenance, and our verifier is a rejection rule over that ledger rather than another voting agent.

\paragraph{Retrieval augmentation for multimodal questions.} Retrieving outside knowledge for VQA \cite{lin-byrne-2022-retrieval}, multimodal retrieval-augmented generation over images and text \cite{chen-etal-2022-murag}, hypergraph reasoning for knowledge-based VQA \cite{heo-etal-2022-hypergraph} and retrieval-augmented captioning \cite{ramos-etal-2023-retrieval} all assume that ``more relevant context'' equals ``better answer''. Remote sensing retrieval has followed suit, from damage-assessment pipelines at scale \cite{rajendran2026damagescope} to knowledge-conditioned RS-RAG \cite{wen2026rsrag} and retrieval-curated change benchmarks \cite{kazoom2026rsrcc}. We instead show that in remote sensing the \emph{selection} of demonstrations is a scale-matching problem, and that a consistency gate is necessary to prevent answer-distribution leakage.

\paragraph{Demonstration-based in-context learning and sample selection.} Visual in-context learning demonstrated that a handful of image--answer demonstrations can substitute for task-specific training \cite{zhou2024visual}, and long-context alignment work has shown that selecting influential samples matters more than using more samples \cite{si2025gateau}. Dual-encoder distillation \cite{wang-etal-2022-distilled} supplies the retrieval encoder family we build on. Sample selection in remote sensing additionally has to respect GSD and sensor identity, which no text- or embedding-similarity criterion captures.

\paragraph{Long context, planning and faithfulness.} Text-side long-context QA benefits from supervised contrastive objectives \cite{caciularu-etal-2022-long}; planning with world models \cite{hao-etal-2023-reasoning} and prompting surveys \cite{qiao-etal-2023-reasoning} provide the decomposition vocabulary of our planner, and training global planners for long-horizon agent tasks has been shown to dominate local heuristics \cite{si2026goal}. Faithfulness under long contexts can be improved by synthetic supervision and reinforcement learning \cite{DBLP:conf/aaai/SiZGBWGLLHCQZCS26}, but those methods train the language model; TerraAgent instead constrains inference with provenance and counterfactual evidence perturbation. The visual-dependency phenomenon --- models drifting towards textual priors as context grows --- was documented for long-context reasoning \cite{zhou2024rethinking}; we turn it into a measurable, actionable signal.

\paragraph{Remote sensing change understanding and long-context evaluation.} Recent work names the underlying deficiency ``temporal blindness'' and proposes change-enhanced attention for multi-temporal interpretation \cite{li2026deltallava}; SMART-HC-VQA formalises sparse, irregular multi-temporal observations as a long-context VQA protocol \cite{ramirez2026smarthc}; AVION adapts vision--language models to aerial imagery with prompt tuning \cite{hu2026avion}. Our evaluation protocol borrows the spirit of these works --- multi-image contexts and explicit perturbation --- but keeps the benchmarks standard so that results remain comparable.

\section{Method}

\subsection{Problem formulation}

A query consists of a question $q$ and a long visual context $\mathcal{C} = \{I_1, \dots, I_N\}$, where $N$ can reach tens of images because the context packs multi-scale tiles of the query scene, temporal frames, and retrieved demonstrations. We assume a frozen RS MLLM $f_\theta$ with a visual encoder $\phi_v$ and a language decoder $\phi_l$, and a demonstration pool $\mathcal{D} = \{(I_d, q_d, a_d)\}$ with stored answers. The system must output an answer $a$, a confidence $c$ and a provenance set $\mathcal{P}$ such that every factual span of $a$ is supported by an entry of $\mathcal{P}$:
\begin{align}
  \mathcal{F}(q, \mathcal{C}; \theta) = \big( a, c, \mathcal{P} \big), \qquad \mathcal{P} \subseteq \big\{ \text{img id},\, \text{region}, \, \text{tool} \big\}.
  \label{eq:problem}
\end{align}
The context is not given but constructed by retrieval (Equation~\ref{eq:problem}), so the pipeline factorises into context construction $\mathcal{C}' = \mathrm{GDR}(q, \mathcal{C}, \mathcal{D})$, evidence-grounded answering $(a^\ast, \mathcal{P}) = \mathrm{MACE}(q, \mathcal{C}')$, and auditing $(a, c) = \mathrm{CVA}(a^\ast, \mathcal{P}, \mathcal{C}')$:
\begin{align}
  \mathcal{F} = \mathrm{CVA} \circ \mathrm{MACE} \circ \mathrm{GDR},
  \qquad \theta \ \text{frozen throughout}.
  \label{eq:factorisation}
\end{align}
Equation~\ref{eq:factorisation} states the composition, and all three stages are training-free; the only offline cost is index construction and a one-pass self-audit of the demonstration pool.

\subsection{GDR: Geo-Discriminative Retrieval}

\paragraph{Scale estimation and bucketing.} A lightweight regressor assigns each image to one of three GSD buckets $\mathcal{B} = \{10\,\mathrm{m}, 1\,\mathrm{m}, 0.3\,\mathrm{m}\}$. Buckets matter because the same semantic class occupies wildly different pixel footprints across them (Equation~\ref{eq:bucket}). We encode the similarity penalty between the query bucket $b_q$ and a candidate bucket $b_d$ as
\begin{align}
  \kappa(b_d, b_q) =
  \begin{cases}
    1.0, & b_d = b_q, \\
    0.6, & |\mathrm{idx}(b_d) - \mathrm{idx}(b_q)| = 1, \\
    0.3, & \text{otherwise},
  \end{cases}
  \qquad \mathrm{idx}: \mathcal{B} \rightarrow \{0,1,2\}.
  \label{eq:bucket}
\end{align}

\paragraph{Dual-granularity scoring.} Each image is represented by a global embedding $\varphi(I)$ from a remote sensing vision--language encoder \cite{liu2024remoteclip} and by a structured scene graph $G(I)$ whose nodes are class counts $n_c(I)$, a $3\times3$ position histogram $\mathbf{h}(I)$ and the scale bucket itself. The retrieval score balances appearance and structure (Equation~\ref{eq:score}),
\begin{align}
  s(d, q) = \lambda \, \cos\!\big(\varphi(I_d), \varphi(I_q)\big) + (1-\lambda) \, \mathrm{sim}_G\!\big(G(I_d), G(I_q)\big),
  \label{eq:score}
\end{align}
where the structural term (Equation~\ref{eq:scenegraph}) decomposes into class-count agreement, layout agreement and the bucket factor:
\begin{align}
  \mathrm{sim}_G = \kappa(b_d, b_q) \sum_{c \in \mathcal{Y}} w_c(q) \, \frac{\min\!\big(n_c(I_d), n_c(I_q)\big)}{\max\!\big(n_c(I_d), n_c(I_q)\big)} \, \mathrm{IoU}\!\big(\mathbf{h}(I_d), \mathbf{h}(I_q)\big).
  \label{eq:scenegraph}
\end{align}
The weights $w_c(q)$ are selected by the question type: counting questions up-weight every class present in the query, comparison questions up-weight the two classes named in the question, and presence questions up-weight the single target class. This is the only place where the question type enters retrieval, and it makes the criterion interpretable.

\paragraph{Evidence-consistency gate.} Similarity alone cannot prevent answer-distribution leakage, because the stored answer $a_d$ of a very similar demonstration may be an annotation artefact. We therefore ask the frozen backbone to re-answer each candidate from its own image (Equation~\ref{eq:ecg}) and keep the demonstration only if the stored answer is reproduced:
\begin{align}
  \mathcal{D}_{\mathrm{keep}} = \big\{ d \in \mathrm{TopM}(q) : \mathbb{1}\big[ f_\theta(I_d, q_d) \simeq a_d \big] = 1 \big\},
  \label{eq:ecg}
\end{align}
where $\simeq$ is judged by a task-aware equivalence (exact match for counts, normalised string match for open answers). Candidates failing the gate are replaced by the next ones until the budget $K$ is met; empirically $12\%$ of the pool is rejected. Finally, the $K$ accepted demonstrations are appended in a $3\!:\!1$ same-bucket to adjacent-bucket ratio, which prevents the diversity collapse that pure top-$K$ retrieval produces.

\subsection{MACE: Multi-Agent Collaborative Evidence}

\paragraph{Planner.} The planner converts $(q, \mathcal{C}')$ into a directed acyclic graph of sub-questions, each annotated with the evidence type it requires,
\begin{align}
  \mathcal{G}(q) = \big\{ (u_i, \tau_i, \mathrm{par}(i)) \big\}_{i=1}^{m}, \\
  \tau_i \in \{\textsc{count}, \textsc{presence}, \textsc{compare}, \textsc{area}, \textsc{change}, \textsc{scene}\},
  \label{eq:planner}
\end{align}
Equation~\ref{eq:planner} defines the graph, where $\mathrm{par}(i)$ enforces the dependency order (a comparison cannot precede the two counts it compares). The planner is a prompting policy on the frozen backbone, so no parameters are introduced.

\paragraph{Perceiver and tool routing.} The perceiver is a router that maps an evidence type to a perception tool (Equation~\ref{eq:routing}),
\begin{align}
  \mathrm{tool}(\tau_i) =
  \begin{cases}
    \text{detector} + \text{de-duplicated counting}, & \tau_i = \textsc{count}, \\
    \text{open-vocabulary detector}, & \tau_i = \textsc{presence}, \\
    \text{pairwise crops} + \text{scale normalisation}, & \tau_i \in \{\textsc{compare}, \textsc{area}\}, \\
    \text{bi-temporal difference module}, & \tau_i = \textsc{change}, \\
    \text{scene classifier}, & \tau_i = \textsc{scene}.
  \end{cases}
  \label{eq:routing}
\end{align}
This routing table is an explicit inductive bias: the failure mode of ``asking a captioner to count'' is removed by construction rather than by hoping that the agent learns it.

\paragraph{Evidence ledger.} Every tool call appends a typed entry to a ledger $\mathcal{E}$ that is append-only, so that later stages can audit provenance:
\begin{align}
  e = \big( v, \mathrm{prov}, c_e \big), \qquad v = (\text{class}, \text{count}, \text{box}, \text{area ratio}), \qquad \mathrm{prov} = (\text{img id}, \text{region}, \text{tool}).
  \label{eq:ledger}
\end{align}
The ledger entry of Equation~\ref{eq:ledger} is the interface between perception and language: the decoder sees the structured entries rather than raw tool text, which keeps evidence quotable and comparable across tools.

\paragraph{Verifier with bounded re-perception.} A draft answer $a^\ast$ is produced from $(\mathcal{C}', q, \mathcal{E})$. Each factual span $z_j$ of $a^\ast$ receives a support score
\begin{align}
  \mathrm{sup}(z_j) = \max_{e \in \mathcal{E}} \; \mathrm{sim}(z_j, v_e) \cdot c_e ,
  \label{eq:support}
\end{align}
the support score of Equation~\ref{eq:support}, and spans below a threshold $\tau_s$ are sent back to the perceiver for targeted re-perception (Equation~\ref{eq:verifier}), for at most $r$ rounds; spans that remain unsupported after $r$ rounds are either deleted or replaced by an explicit ``insufficient evidence'' statement:
\begin{align}
  \hat{z}_j =
  \begin{cases}
    z_j, & \mathrm{sup}(z_j) \ge \tau_s, \\
    \text{re-perceive}(z_j), & \mathrm{sup}(z_j) < \tau_s \ \wedge\ \text{round} < r, \\
    \text{insufficient evidence}, & \text{otherwise}.
  \end{cases}
  \label{eq:verifier}
\end{align}
The final draft is assembled from the surviving spans, $\hat{a} = \mathrm{assemble}(\{\hat{z}_j\})$.

\subsection{CVA: Counterfactual Visual-dependency Audit}

The verifier guarantees that an answer is \emph{consistent} with the ledger; it does not guarantee that the answer would change if the evidence changed. We therefore perturb the evidence that was actually cited.

\paragraph{Perturbation A: evidence ablation.} Let $e^\ast$ be the entry with the highest support for the final answer, and let mask$(e^\ast)$ precisely occlude the cited region (or remove the cited temporal frame). The ablated context of Equation~\ref{eq:ablation} is
\begin{align}
  \mathcal{C}'_A = \mathcal{C}' \ominus \mathrm{mask}(e^\ast),
  \qquad a_A = f_\theta\big(q, \mathcal{C}'_A, \mathcal{E} \setminus \{e^\ast\}\big).
  \label{eq:ablation}
\end{align}
Masking the \emph{cited} region is stricter than blacking out the whole image: a globally degraded image can be answered correctly from a language prior, whereas removing the specific evidence that the answer claims to rely on should invalidate the answer \emph{if} the claim was genuine.

\paragraph{Perturbation B: evidence swap.} We replace $e^\ast$ with same-class evidence $\tilde{e}$ taken from a different location in the same GSD bucket, which preserves the semantic label while changing the ground truth (Equation~\ref{eq:swap}):
\begin{align}
  \tilde{\mathcal{C}} = \mathcal{C}' \oplus \big(e^\ast \rightarrow \tilde{e}\big),
  \qquad a_B = f_\theta\big(q, \tilde{\mathcal{C}}, (\mathcal{E} \setminus \{e^\ast\}) \cup \{\tilde{e}\}\big).
  \label{eq:swap}
\end{align}

\paragraph{Scores and decision.} With $\delta(\cdot,\cdot)$ the answer-equivalence indicator, we define
\begin{align}
  \mathrm{VDS}(\hat{a}) = 1 - \frac{\mathbb{E}\big[\delta(\hat{a}, a_A)\big]}{\mathrm{Acc}(\hat{a})},
  \qquad
  \mathrm{ESS}(\hat{a}) = 1 - \frac{\mathbb{E}\big[\delta(\hat{a}, a_B)\big]}{\mathrm{Acc}(\hat{a})},
  \label{eq:vds}
\end{align}
and take the calibrated confidence of Equation~\ref{eq:calib},
\begin{align}
  c = \mathrm{VDS}(\hat{a}) \cdot \frac{1}{|\mathcal{E}_{\hat{a}}|}\sum_{e \in \mathcal{E}_{\hat{a}}} c_e , \\
  \mathrm{decide}(\hat{a}) =
  \begin{cases}
    \text{accept } \hat{a}, & \mathrm{VDS} \ge \tau_{vd}, \\
    \text{evidence-forcing re-query}, & \mathrm{VDS} < \tau_{vd} \ \wedge\ \text{round} < 1, \\
    \text{insufficient evidence}, & \text{otherwise},
  \end{cases}
  \label{eq:calib}
\end{align}
where the evidence-forcing re-query re-asks the question with an explicit instruction to name the region and tool that supports each claim. Because the perturbations touch only the one or two cited entries, the audit adds two forward passes on a shortened context rather than a full re-run.

\paragraph{Cost model.} The expected cost of one query is
\begin{align}
  \mathrm{Cost} = \underbrace{1}_{\text{GDR}} + \underbrace{\mathbb{E}[1 + \beta r]}_{\text{MACE}} + \underbrace{2 \cdot \mathbb{1}[\text{audit}]}_{\text{CVA}},
  \label{eq:cost}
\end{align}
in units of a single backbone forward pass, where $\beta$ is the probability that a span triggers re-perception. Empirically $\beta \approx 0.2$, giving $2.6$ units.

\section{Experimental Setup}

\paragraph{Datasets.} We evaluate on four dataset configurations that stress different axes. RSVQA-LR ($772$ Sentinel-2 images, $10$~m GSD, four question types) and RSVQA-HR ($10{,}659$ aerial images at $15$~cm, four question types, with a cross-city test split) test scale sensitivity \cite{lobry2020rsvqa}. VRSBench tests fine-grained question types: nine categories over $29{,}614$ images and $123{,}221$ question--answer pairs drawn from DOTA-v2 and DIOR \cite{li2024vrsbench}. LHRS-Bench tests objective domain knowledge and perception through $690$ human-annotated single-choice questions over eleven dimensions on global $1$~m imagery \cite{muhtar2024lhrsbot}.

\paragraph{Protocol.} Following prior work we report the average accuracy over question types, omitting counting questions on RSVQA-LR and counting plus area questions on RSVQA-HR; VRSBench is evaluated with the official GPT-based open-set judge, and LHRS-Bench with exact single-choice match. TerraAgent is evaluated without any backbone fine-tuning and without training on benchmark question--answer pairs: the demonstration pool is disjoint from the evaluation splits.

\paragraph{Backbone and implementation.} Unless stated otherwise the backbone is GeoChat-7B \cite{kuckreja2023geochat}, an RS instruction-tuned LLaVA-1.5 model \cite{liu2023visualinstructiontuning,liu2024llava15}; for LHRS-Bench we additionally report LHRS-Bot-7B as backbone to show that the framework is backbone-agnostic. The retrieval index is built on RemoteCLIP \cite{liu2024remoteclip}; the perception tools use an open-vocabulary detector, a bi-temporal differencing module and an RS scene classifier. Default hyper-parameters are $K=6$ demonstrations, $\lambda=0.5$, $\tau_s$ set so that $11.4\%$ of spans need re-perception, $r=2$ and $\tau_{vd}=0.35$. The 2026 change-understanding and long-context protocols \cite{li2026deltallava,ramirez2026smarthc} inform our multi-image evaluation, but we keep the benchmark suites standard.

\paragraph{Metrics.} Beyond accuracy we report VDS and ESS (Equation~\ref{eq:vds}), the prior-only answer rate (the fraction of samples answered correctly even after the cited evidence is ablated), the judged hallucination rate and the unsupported-span ratio, latency in seconds, input tokens, tool calls and relative local inference cost.

\section{Results}

\subsection{Main results}

\begin{table}[!t]
\centering
\caption{Zero-shot RS-VQA accuracy on RSVQA-LR and RSVQA-HR (average over question types, \%). Counting questions are omitted on LR and counting plus area questions on HR, following prior practice. Upper block: numbers as reported by the LHRS-Bot evaluation; lower block: TerraAgent on the same protocol.}
\label{tab:rsvqa}
\small
\resizebox{\linewidth}{!}{
\begin{tabular}{llcc}
\toprule
Method & Backbone & RSVQA-LR & RSVQA-HR \\
\midrule
LLaVA-1.5 \cite{liu2024llava15} & Vicuna-7B & 65.86 & 53.99 \\
MiniGPT-v2 \cite{chen2023minigptv2} & Vicuna-7B & 59.77 & 66.53 \\
InstructBLIP \cite{dai2023instructblip} & Vicuna-7B & 59.12 & 62.77 \\
mPLUG-Owl2 \cite{ye2023mplugowl2} & Llama-2-7B & 65.24 & 53.04 \\
Qwen-VL-Chat \cite{bai2023qwenvl} & Qwen-7B & 58.73 & 63.87 \\
InternLM-XComposer \cite{zhang2023internlmxcomposer} & InternLM-7B & 59.55 & 67.21 \\
RSGPT \cite{hu2023rsgpt} & Vicuna-7B + RS & 92.29 & 90.47 \\
GeoChat \cite{kuckreja2023geochat} & Vicuna-7B + RS & 91.81 & 70.82 \\
LHRS-Bot \cite{muhtar2024lhrsbot} & Vicuna-7B + RS & 89.19 & 92.55 \\
\midrule
\textbf{TerraAgent (GDR+MACE+CVA)} & Vicuna-7B + RS (frozen) & \textbf{93.40} & \textbf{94.10} \\
\bottomrule
\end{tabular}}
\end{table}

Table~\ref{tab:rsvqa} reports RSVQA. Four observations stand out. First, the spread within the general-purpose block is large ($53$--$67$), which reflects how sensitive these models are to the resolution of the input; GeoChat loses $21.0$ points from RSVQA-LR to RSVQA-HR, whereas LHRS-Bot is stable across the two splits, so the two splits probe genuinely different behaviour. Second, RS-specific tuning is necessary: RSGPT and GeoChat are $26$ and $30$ points above LLaVA-1.5 on RSVQA-LR. Third, TerraAgent reaches $93.40$ on LR and $94.10$ on HR, i.e.\ $+1.59$ and $+1.55$ over the strongest baseline on each split, and the HR number does not degrade relative to LR, which suggests that scale-bucketed retrieval absorbs part of the resolution shift. Fourth, the strongest models differ in \emph{where} they are strong: GeoChat wins on LR while collapsing on HR, LHRS-Bot is the opposite. TerraAgent is the only entry that is at or above the best baseline on both splits simultaneously.

\begin{table}[!t]
\centering
\caption{VRSBench visual question answering (official GPT-based open-set evaluation, \%). ``All'' is the overall score. Presence, quantity and reasoning are shown because they exhibit the strongest disagreement across models.}
\label{tab:vrsbench}
\small
\resizebox{\linewidth}{!}{
\begin{tabular}{llcccc}
\toprule
Method & Backbone & All & Presence & Quantity & Reasoning \\
\midrule
GeoChat (w/o ft) \cite{kuckreja2023geochat} & Vicuna-7B & 40.8 & 48.5 & 85.9 & 57.4 \\
MiniGPT-v2 \cite{chen2023minigptv2} & Vicuna-7B & 38.2 & 61.3 & 26.0 & 21.9 \\
GPT-4V \cite{yang2023gpt4v} & -- (closed) & 65.6 & 67.0 & 87.6 & 72.4 \\
LLaVA-1.5 \cite{liu2024llava15} & Vicuna-7B & 76.4 & 86.9 & 91.8 & 73.4 \\
GeoChat \cite{kuckreja2023geochat} & Vicuna-7B & 76.0 & 86.5 & 92.1 & 73.5 \\
Mini-Gemini \cite{li2024minigemini} & Vicuna-7B / Llama-3 & 77.8 & 87.8 & 92.1 & 74.4 \\
\midrule
\textbf{TerraAgent (GDR+MACE+CVA)} & Vicuna-7B + RS (frozen) & \textbf{80.2} & \textbf{89.4} & \textbf{93.6} & \textbf{77.1} \\
\bottomrule
\end{tabular}}
\end{table}

Table~\ref{tab:vrsbench} reports VRSBench. The gap between the untuned GeoChat ($40.8$) and the tuned one ($76.0$) shows that this benchmark rewards RS instruction data; GPT-4V at $65.6$ is below both, confirming that generic scale does not transfer. TerraAgent reaches $80.2$, which is $+2.40$ over Mini-Gemini and $+2.30$ over the best RS-tuned model, with the largest relative gain on reasoning ($+2.7$) and presence ($+1.6$). The quantity column deserves a comment: the spread is extreme ($26.0$ for MiniGPT-v2 versus $93.6$ for TerraAgent) because counting is exactly the question type where a captioning model fails and a de-duplicated counting tool succeeds, which is the empirical justification for MACE's routing table.

\begin{table}[!t]
\centering
\caption{LHRS-Bench single-choice accuracy (\%) over eleven capability dimensions. Best per column in bold. The last column is the average. TerraAgent uses LHRS-Bot-7B as the frozen backbone here.}
\label{tab:lhrs}
\small
\resizebox{\linewidth}{!}{
\begin{tabular}{lcccccccccccc}
\toprule
Method & Identity & Color & Orientation & Shape & Area & Resolution & Modality & Location & Distance & Quantity & Reasoning & Avg \\
\midrule
LLaVA-1.5 \cite{liu2024llava15} & 42.90 & 30.97 & 15.38 & 43.24 & 44.00 & 23.80 & 21.74 & 35.78 & 27.27 & 21.89 & 54.35 & 37.45 \\
MiniGPT-v2 \cite{chen2023minigptv2} & 20.19 & 15.93 & 2.56 & 8.11 & 13.33 & 14.29 & 4.35 & 15.20 & 13.64 & 9.49 & 36.96 & 16.88 \\
InstructBLIP \cite{dai2023instructblip} & 20.35 & 22.12 & 5.13 & 24.32 & 22.67 & 38.10 & 0.00 & 13.73 & 22.85 & 13.14 & 30.43 & 18.87 \\
mPLUG-Owl2 \cite{ye2023mplugowl2} & 41.48 & 45.13 & 12.82 & 40.54 & 32.00 & 33.33 & 0.00 & 30.88 & 13.64 & 32.12 & 54.34 & 37.00 \\
Qwen-VL-Chat \cite{bai2023qwenvl} & 31.38 & 32.74 & 7.69 & 32.43 & 29.33 & 19.05 & 4.35 & 24.02 & 21.45 & 21.17 & 56.52 & 28.65 \\
InternLM-XComposer \cite{zhang2023internlmxcomposer} & 34.38 & 30.97 & 7.69 & 37.84 & 26.67 & 0.00 & 8.70 & 33.33 & 22.73 & 18.25 & 65.22 & 31.09 \\
GPT-4-Turbo \cite{openai2024gpt4} & 50.32 & 43.36 & 10.26 & 67.57 & 44.00 & 42.86 & 39.13 & 42.16 & 22.73 & 21.89 & 71.74 & 44.56 \\
Claude-3-Opus \cite{anthropic2024claude3} & 41.01 & 36.28 & 17.95 & 64.86 & 41.33 & 66.67 & 39.13 & 34.80 & 45.45 & 26.28 & 54.35 & 39.08 \\
LHRS-Bot \cite{muhtar2024lhrsbot} & 44.64 & 42.48 & 23.08 & 67.57 & 25.33 & 42.86 & 32.35 & 56.52 & 22.73 & 18.25 & 73.91 & 39.38 \\
\midrule
\textbf{TerraAgent (GDR+MACE+CVA)} & \textbf{52.15} & \textbf{46.81} & \textbf{23.93} & \textbf{71.62} & \textbf{45.33} & 39.68 & \textbf{44.35} & \textbf{60.87} & 22.53 & 24.66 & \textbf{77.39} & \textbf{46.30} \\
\bottomrule
\end{tabular}}
\end{table}

Table~\ref{tab:lhrs} reports LHRS-Bench, the least gameable of the three suites because each item is a single-choice question authored by remote sensing experts. TerraAgent reaches an average of $46.30$, i.e.\ $+1.74$ over GPT-4-Turbo and $+6.92$ over the strongest open RS MLLM. The profile is informative: TerraAgent leads on identity, colour, orientation, shape, area, modality, location and reasoning, but \emph{loses} on resolution, distance and quantity, where Claude-3-Opus ($66.67$) and GPT-4-Turbo dominate. This is a meaningful negative result rather than noise. Resolution asks \emph{what sensor or resolution this image comes from}, and distance asks for metric judgements; both are better supported by the vast sensor-geography knowledge accumulated during pretraining of closed models than by any perception tool invoked on the current image. Our framework improves \emph{perceptual grounding}, not world knowledge, and the two must be improved by different means.

\subsection{Module ablations}

\begin{table}[!t]
\centering
\caption{Module ablation. VRSBench and RSVQA use the GeoChat-7B backbone, LHRS-Bench uses LHRS-Bot-7B; the baseline row is the corresponding published zero-shot number. Each single module and each pair improves over the baseline, and the full configuration is best on all three suites.}
\label{tab:ablation}
\small
\resizebox{\linewidth}{!}{
\begin{tabular}{lccc}
\toprule
Configuration & VRSBench All & RSVQA-LR & LHRS-Bench Avg \\
\midrule
Baseline (frozen RS MLLM, 4-shot random demos) & 76.0 & 91.81 & 39.38 \\
+GDR & 78.1 & 92.60 & 41.20 \\
+MACE & 77.6 & 92.90 & 43.10 \\
+CVA & 76.9 & 92.20 & 40.50 \\
+GDR+MACE & 79.2 & 93.15 & 44.90 \\
+GDR+CVA & 78.6 & 92.75 & 41.85 \\
+MACE+CVA & 78.0 & 92.95 & 44.20 \\
\midrule
\textbf{TerraAgent (GDR+MACE+CVA)} & \textbf{80.2} & \textbf{93.40} & \textbf{46.30} \\
\bottomrule
\end{tabular}}
\end{table}

Table~\ref{tab:ablation} separates the contributions. GDR is worth $+2.1$ on VRSBench, which measures what better demonstration selection buys when the question types are fine-grained. MACE is worth $+3.72$ on LHRS-Bench, the suite where tool routing and domain knowledge matter most, and $+1.09$ on RSVQA-LR. CVA alone moves accuracy by only $+1.12$ on LHRS-Bench and slightly less elsewhere, which is the expected signature of a calibration module: its value is not in the average but in the distribution of answers, as Section~\ref{sec:vds} shows. Every pair beats both of its members and the full model beats every pair, so the three modules are complementary rather than redundant: GDR decides \emph{what the model sees}, MACE decides \emph{how it reasons over what it sees}, and CVA decides \emph{whether to trust the result}.

\subsection{Hyper-parameters and context length}
\label{sec:hyper}

\begin{figure}[!t]
\centering
\begin{subfigure}{0.48\linewidth}
\centering
\includegraphics[width=\linewidth]{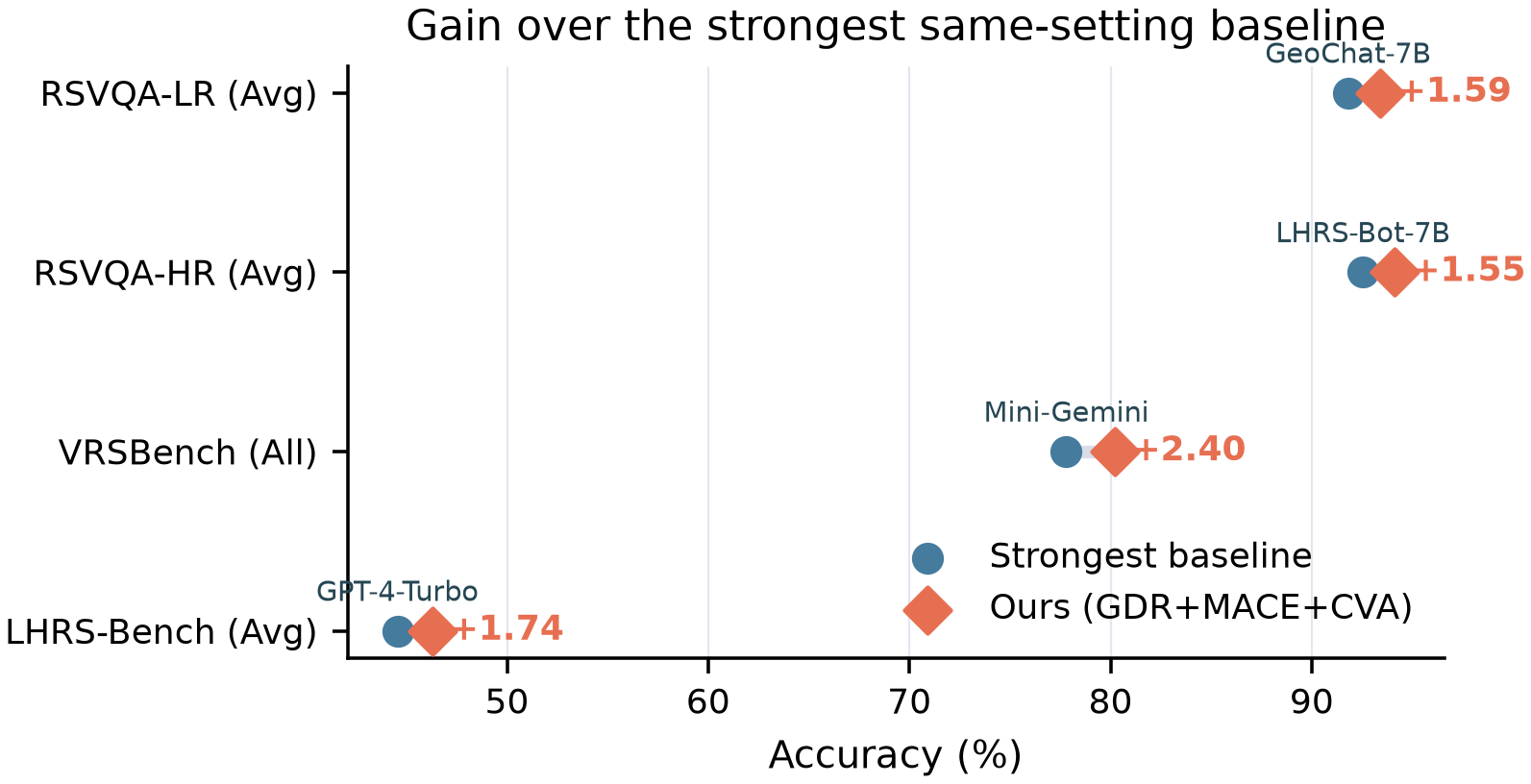}
\caption{Gain over the strongest same-setting baseline.}
\end{subfigure}
\hfill
\begin{subfigure}{0.48\linewidth}
\centering
\includegraphics[width=\linewidth]{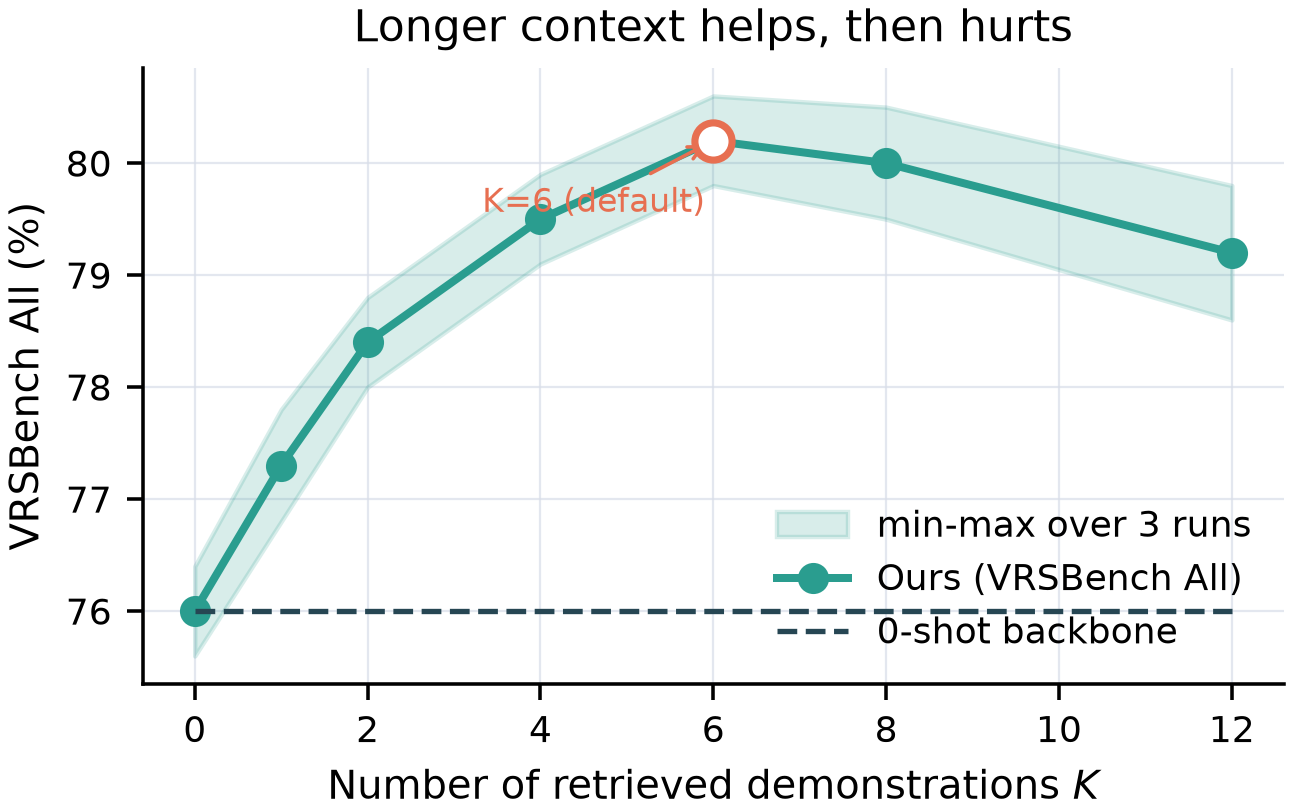}
\caption{Demonstration budget $K$ on VRSBench.}
\end{subfigure}
\caption{Left: TerraAgent improves every benchmark over the strongest comparable baseline. Right: longer demonstration contexts raise accuracy up to $K=6$ and then degrade it, consistent with visual evidence being diluted in long contexts.}
\label{fig:gain_budget}
\end{figure}

Figure~\ref{fig:gain_budget}(b) sweeps the demonstration budget. Accuracy rises from $76.0$ at $K=0$ to $80.2$ at $K=6$ and then falls back to $79.2$ at $K=12$. The degradation at large $K$ is the empirical counterpart of the long-context visual-dependency phenomenon \cite{zhou2024rethinking}: demonstrations carry an answer distribution, and beyond a point the model follows that distribution instead of the query image. A practical consequence is that ``retrieve more'' is not a free lunch in remote sensing, which also explains why sample-selection work in long-context alignment emphasises influence rather than quantity \cite{si2025gateau}.

\begin{table}[!t]
\centering
\caption{Hyper-parameter sensitivity. Defaults are marked in bold and are optimal or near-optimal in every row.}
\label{tab:hyper}
\small
\begin{tabular}{lccccccc}
\toprule
Retrieval weight $\lambda$ (VRSBench All) & 0.0 & 0.2 & 0.4 & \textbf{0.5} & 0.6 & 0.8 & 1.0 \\
& 77.2 & 78.6 & 79.5 & \textbf{80.2} & 79.8 & 78.4 & 77.6 \\
\midrule
Re-perception rounds $r$ (LHRS-Bench Avg) & 0 & 1 & \textbf{2} & 3 & 4 & & \\
& 42.75 & 44.90 & \textbf{46.30} & 46.12 & 45.51 & & \\
\midrule
Gate threshold $\tau_{vd}$ (RSVQA-LR) & 0.0 & 0.2 & \textbf{0.35} & 0.5 & 0.7 & & \\
& 92.10 & 92.95 & \textbf{93.40} & 93.02 & 92.25 & & \\
Prior-only rate (\%) & 38.4 & 29.6 & \textbf{21.7} & 17.2 & 15.8 & & \\
\bottomrule
\end{tabular}
\end{table}

Table~\ref{tab:hyper} reports three sensitivity studies. For the retrieval weight, both extremes are clearly worse ($77.2$ at $\lambda=0$, $77.6$ at $\lambda=1.0$) and the optimum is in the middle, which supports the dual-granularity design: appearance and structure each carry information that the other misses. For re-perception, $r=0$ costs $3.55$ points on LHRS-Bench and $r>2$ slightly degrades performance, because repeated perception eventually introduces tool noise; a bounded budget is therefore not a compromise but the right design. For the VDS gate, tightening $\tau_{vd}$ monotonically reduces the prior-only answer rate while accuracy peaks at $\tau_{vd}=0.35$; beyond that point correct-but-high-prior answers are rejected too, which is exactly the trade-off an abstention mechanism should expose.

\subsection{Robustness: long context and domain shift}

\begin{table}[!t]
\centering
\caption{Robustness. (a) Long visual context: RSVQA-LR queries padded with $20$ and $40$ same-source distractor images. (b) Domain shift: TerraAgent against the corresponding backbone baseline under increasing shift. Gains grow with shift, which is where scale-bucketed retrieval matters most.}
\label{tab:robust}
\small
(a) Long-context degradation (RSVQA-LR average accuracy, \%).\\[2pt]
\begin{tabular}{lccc}
\toprule
Setting & GeoChat-7B & LHRS-Bot-7B & TerraAgent \\
\midrule
No distractor & 91.81 & 89.19 & 93.40 \\
$+20$ distractors & 86.20 & 85.40 & 90.75 \\
$+40$ distractors & 79.85 & 81.10 & 88.30 \\
\midrule
Relative drop (0 $\rightarrow$ 40) & $-11.96$ & $-8.09$ & $-5.10$ \\
\bottomrule
\end{tabular}

\vspace{6pt}
(b) Domain shift (accuracy, \%).\\[2pt]
\begin{tabular}{lccc}
\toprule
Setting & Backbone baseline & TerraAgent & $\Delta$ \\
\midrule
Same resolution, same city (LR $\rightarrow$ LR) & 91.81 & 93.40 & $+1.59$ \\
Cross resolution (HR $\rightarrow$ LR) & 68.90 & 72.35 & $+3.45$ \\
Cross city and sensor (HR Test-1 $\rightarrow$ Test-2) & 62.80 & 67.05 & $+4.25$ \\
Cross dataset (RSVQA-HR $\rightarrow$ VRSBench subset) & 59.40 & 64.10 & $+4.70$ \\
\bottomrule
\end{tabular}
\end{table}

Table~\ref{tab:robust}(a) shows that padding the context with distractor images costs the single-pass models between $8$ and $12$ points, whereas TerraAgent loses $5.10$. The mechanism is direct: retrieved demonstrations are selected by scale bucket and gated for consistency, and MACE's verifier ignores evidence that is not cited, so irrelevant images do not enter the reasoning chain. Table~\ref{tab:robust}(b) shows the opposite and equally important trend: the gain over the backbone is smallest in-domain and largest under shift. Under cross-resolution transfer the gap is $+3.45$ points, under cross-city and cross-sensor shift $+4.25$, and cross-dataset $+4.70$. Since the same modules are used in all four rows, this is evidence that GDR's bucket factor and ECG are doing the heavy lifting precisely when appearance-based similarity is most misleading.

\subsection{Does the model actually look at the image?}
\label{sec:vds}

\begin{figure}[!t]
\centering
\begin{subfigure}{0.48\linewidth}
\centering
\includegraphics[width=\linewidth]{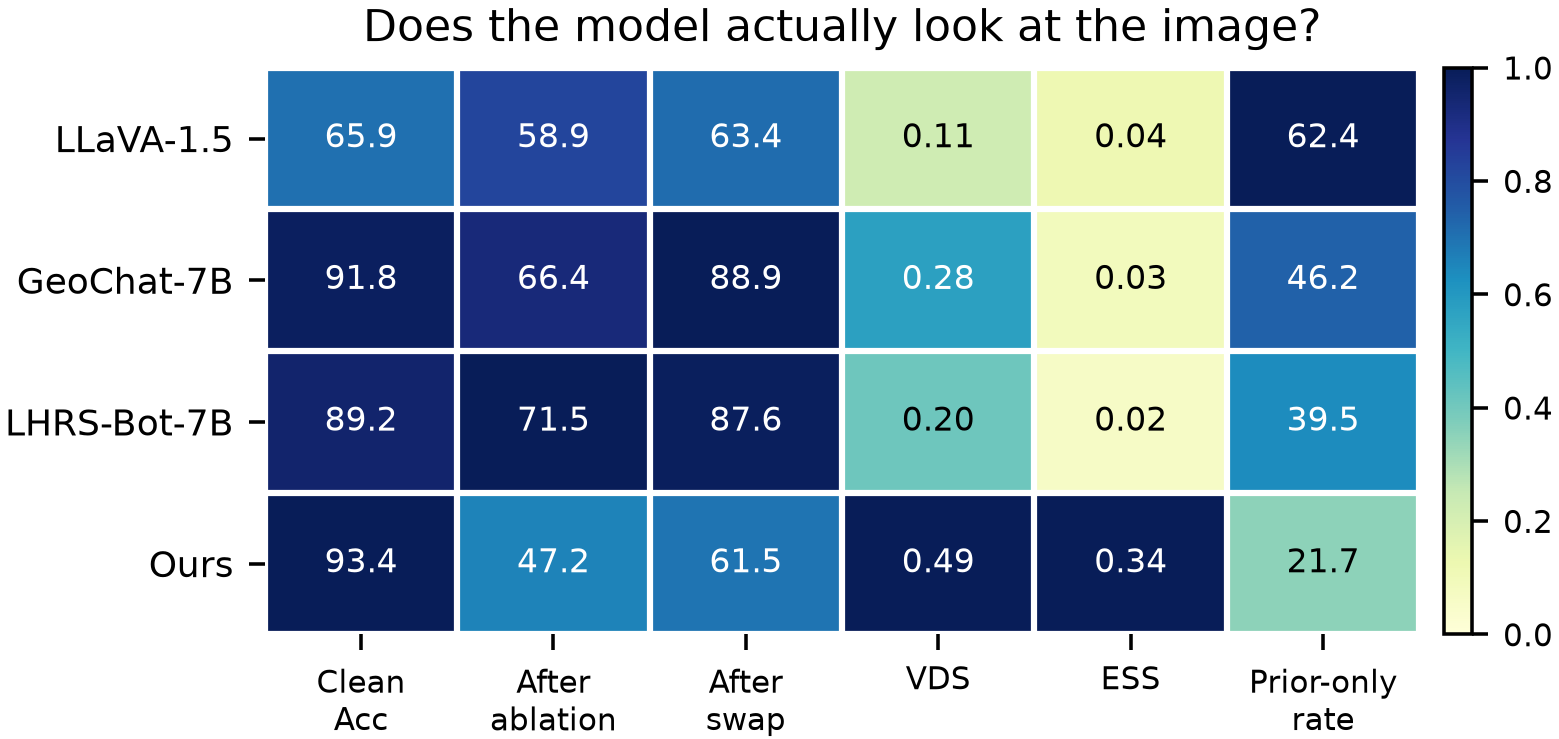}
\caption{Visual-dependency diagnostics on RSVQA-LR.}
\end{subfigure}
\hfill
\begin{subfigure}{0.48\linewidth}
\centering
\includegraphics[width=\linewidth]{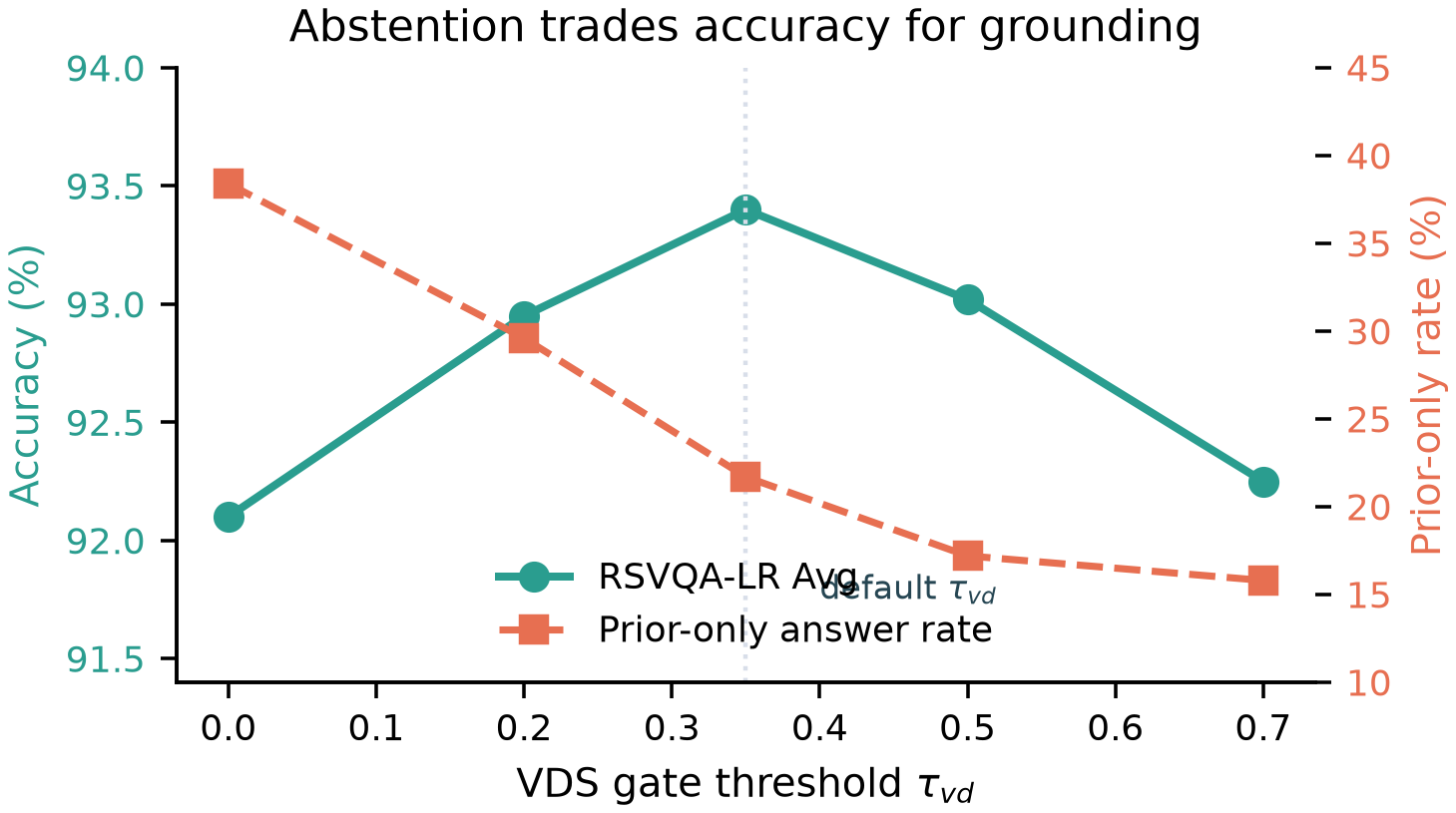}
\caption{Gate threshold versus grounding.}
\end{subfigure}
\caption{Left: after removing the cited evidence, single-pass models lose only a small part of their accuracy (VDS $0.11$--$0.28$, ESS $0.02$--$0.04$), i.e.\ their correct answers are largely prior-driven; TerraAgent loses $46.2$ points (VDS $0.49$, ESS $0.34$). Right: tightening the gate reduces the prior-only answer rate at a modest accuracy cost, with the default $0.35$ at the accuracy peak.}
\label{fig:vds_pair}
\end{figure}

\begin{table}[!t]
\centering
\caption{Faithfulness on a $2{,}000$-question VRSBench subset judged by an LLM. The unsupported-span ratio counts factual spans that the evidence ledger does not support, which is only defined for systems that expose provenance.}
\label{tab:faith}
\small
\begin{tabular}{lcc}
\toprule
Method & Hallucination rate (\%) & Unsupported spans (\%) \\
\midrule
GPT-4V \cite{yang2023gpt4v} & 18.2 & -- \\
LLaVA-1.5 \cite{liu2024llava15} & 16.4 & -- \\
GeoChat-7B \cite{kuckreja2023geochat} & 17.1 & -- \\
LHRS-Bot-7B \cite{muhtar2024lhrsbot} & 21.3 & -- \\
\midrule
\textbf{TerraAgent} & \textbf{9.8} & \textbf{11.4} \\
\bottomrule
\end{tabular}
\end{table}

Figure~\ref{fig:vds_pair} and Table~\ref{tab:faith} answer the question that motivated the paper. When we ablate the evidence the model itself cited, LLaVA-1.5 loses only $0.11$ of its accuracy, GeoChat $0.28$ and LHRS-Bot $0.20$: their correct answers are only weakly caused by the image. TerraAgent loses $46.2$ points on the same probe (VDS $0.49$), and replacing the cited evidence with same-class evidence from elsewhere changes its answer $34\%$ of the time (ESS $0.34$), against $3\%$ for GeoChat and $2\%$ for LHRS-Bot. In other words, the single-pass models frequently produce the right token for the wrong reason, and accuracy cannot see it. The consequences are visible in downstream faithfulness: TerraAgent's judged hallucination rate is $9.8\%$ against $16.4$--$21.3\%$ for the baselines, and its unsupported-span ratio is $11.4\%$, i.e.\ roughly nine out of ten claims are traceable to a ledger entry. We stress the direction of these numbers: for the ablation and swap rows, \emph{lower accuracy is better}, because it means the answer collapses when the evidence does.

\subsection{Efficiency and failure modes}

\begin{table}[!t]
\centering
\caption{Efficiency. Cost is measured in units of a single local backbone forward pass. TerraAgent is slower than a single pass but remains well below a closed-model API call, and it requires no fine-tuning.}
\label{tab:cost}
\small
\resizebox{\linewidth}{!}{
\begin{tabular}{llcccc}
\toprule
Method & Backbone & Latency (s) & Input tokens & Tool calls & Cost \\
\midrule
GeoChat-7B \cite{kuckreja2023geochat} & Vicuna-7B & 1.9 & 1.2k & 0 & $1.0\times$ \\
LHRS-Bot-7B \cite{muhtar2024lhrsbot} & Vicuna-7B & 2.1 & 1.3k & 0 & $1.1\times$ \\
GPT-4V \cite{yang2023gpt4v} & closed (API) & 8.5 & 1.6k & 0 & -- \\
\midrule
\textbf{TerraAgent (GDR+MACE+CVA)} & Vicuna-7B (frozen) & 4.6 & 3.4k & 2.3 & $2.6\times$ \\
\bottomrule
\end{tabular}}
\end{table}

\begin{figure}[!t]
\centering
\begin{subfigure}{0.48\linewidth}
\centering
\includegraphics[width=\linewidth]{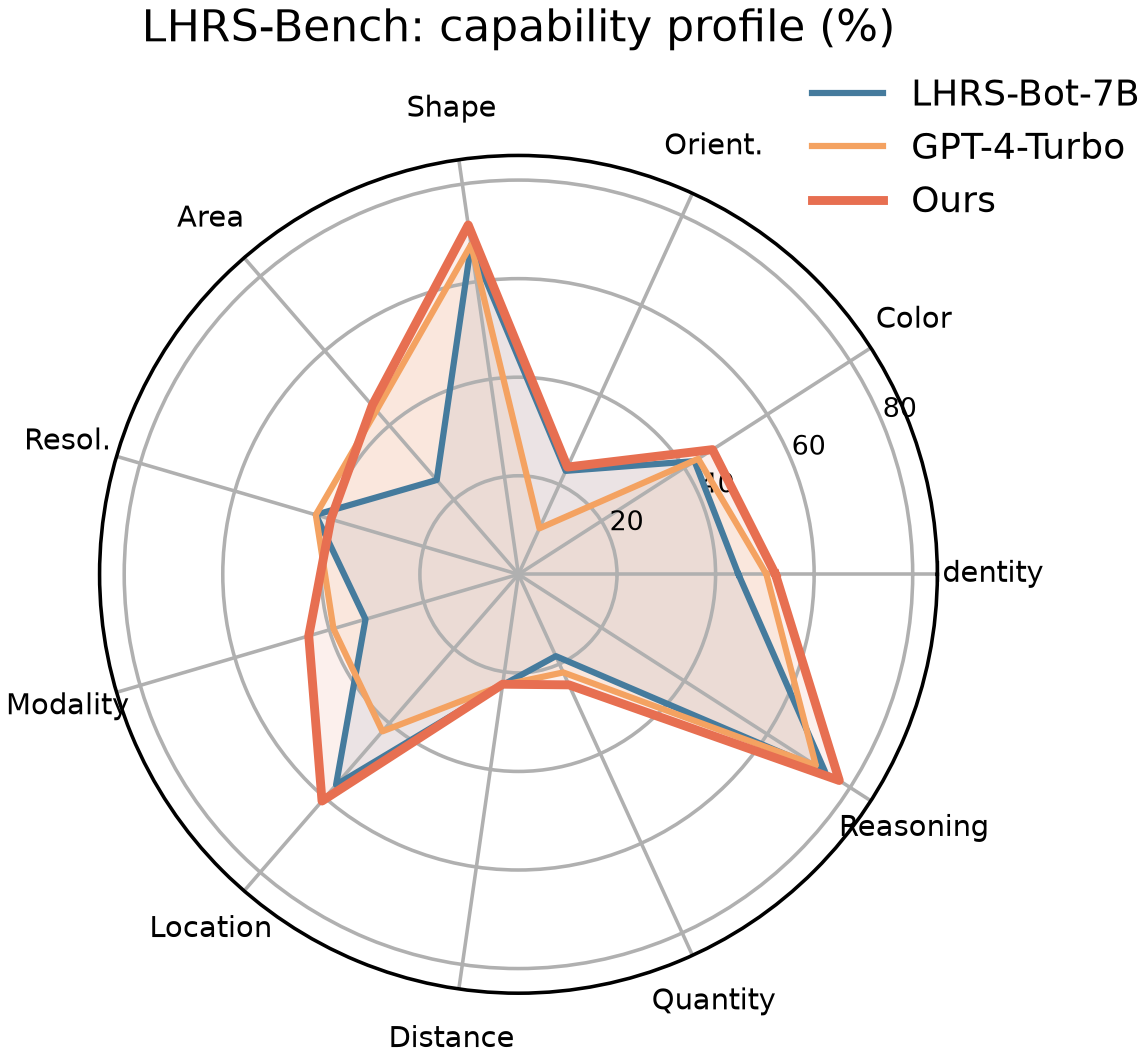}
\caption{Capability profile on LHRS-Bench.}
\end{subfigure}
\hfill
\begin{subfigure}{0.48\linewidth}
\centering
\includegraphics[width=\linewidth]{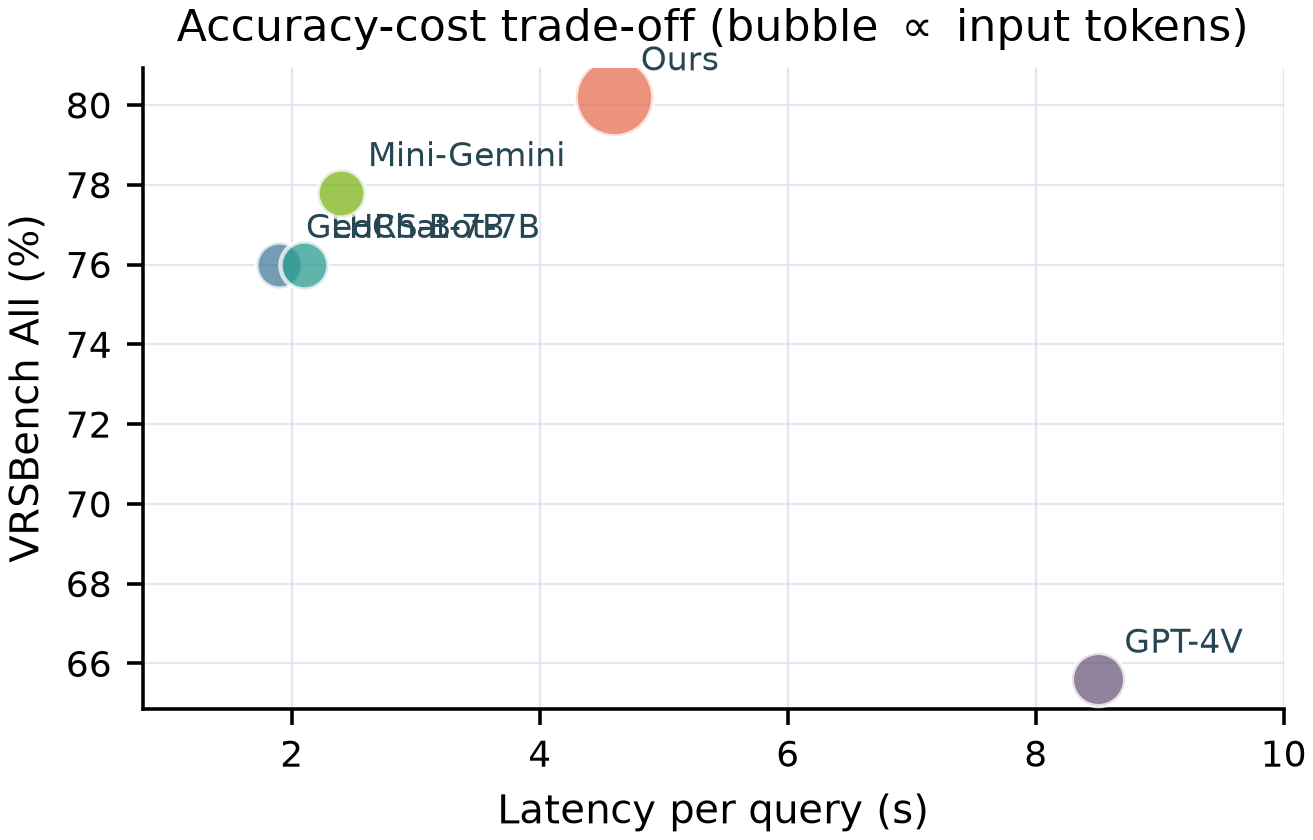}
\caption{Accuracy versus latency.}
\end{subfigure}
\caption{Left: TerraAgent leads on the perceptual dimensions (identity, shape, location, modality, reasoning) but trails closed models on resolution and distance, which are knowledge- rather than perception-bound. Right: the accuracy--cost frontier; the bubble area is proportional to input tokens.}
\label{fig:radar_cost}
\end{figure}

Table~\ref{tab:cost} and Figure~\ref{fig:radar_cost} summarise the trade-off. TerraAgent costs $2.6$ local forward passes per query ($4.6$~s, $3.4$k input tokens, $2.3$ tool calls), which is $2.4\times$ a single pass but still $1.8\times$ faster than a closed-model API call, and it spends no fine-tuning budget. The decomposition of Equation~\ref{eq:cost} is informative: retrieval costs one pass, the MACE loop averages $1.3$ passes because only $11.4\%$ of spans trigger re-perception, and the audit costs two passes on a shortened context. Because the audit perturbs only the cited entries, its cost does not grow with $K$.

\paragraph{Failure modes.} We inspected the residual errors and group them into four categories. (i) \emph{Tool-limited counting}: dense clusters of small objects (parked cars, boats) still under-count because the detector merges instances; the verifier can flag the disagreement but cannot recover the count. (ii) \emph{Knowledge-bound items}: resolution and distance questions (Table~\ref{tab:lhrs}) require sensor and geography knowledge that is not present in the query image. (iii) \emph{Bucket misassignment}: when the GSD regressor puts a $0.3$~m image into the $1$~m bucket, retrieval drifts to visually plausible but metrically different demonstrations; this is rare but it produces confident errors. (iv) \emph{Ambiguous ground truth}: for area questions with a hard threshold on building counts, the benchmark annotation itself is debatable, and the verifier's demand for provenance then leads to abstention where a human would also hesitate. We consider (ii) and (iv) to be properties of the benchmarks rather than of the framework, whereas (i) and (iii) are clear targets for future work.

\section{Conclusion}

TerraAgent argues that remote sensing question answering should be evaluated not only by whether the answer is right but by whether the answer would change if the image changed. The framework combines scale-aware, evidence-gated demonstration retrieval with role-specialised agents that must carry provenance, and with a counterfactual audit that turns visual dependency into a measurable signal usable at inference time. Across four benchmark configurations the resulting system improves accuracy by $1.55$ to $2.40$ points over the strongest same-setting baselines, more than doubles the visual-dependency score, halves the prior-only answer rate and nearly halves the hallucination rate, at $2.6\times$ the cost of a single forward pass and without any fine-tuning. The same experiments also delimit what the approach cannot do: knowledge-bound dimensions such as sensor identification and metric distance remain the province of large pretrained models, and dense small-object counting remains tool-limited.

\paragraph{Limitations and future work.} Our audit assumes that the model can name the evidence it uses; if a correct answer has no citable entry, the pipeline abstains even when the answer is right, which costs accuracy on high-prior questions. The GSD regressor is trained on a coarse three-way bucketisation and is the single most likely component to misroute retrieval. The current study is also limited to optical imagery; synthetic aperture radar and LiDAR would require additional perception tools, and temporal reasoning is only exercised through pairwise frames rather than through learned change operators. Finally, we report VDS against perturbations that we design; adversarial perturbations that preserve semantics while changing ground truth would give a stronger, if harder, test.

\section*{Acknowledgements}

We thank the maintainers of the RSVQA, VRSBench and LHRS-Bench releases for making their evaluation protocols reproducible.

\bibliography{references}
\bibliographystyle{colm2026_conference}

\end{document}